\RequirePackage{fix-cm}
\documentclass[smallextended]{svjour3}       
\smartqed  
\usepackage{graphicx}
 \usepackage{amsmath}
\usepackage{enumerate}

\begin{document}

\title{Image Classification Based on Quantum KNN Algorithm\thanks{This work is supported by the National Natural Science Foundation of China under Grants No. 61502016 and 61771230, the Joint Open Fund of Information Engineering Team in Intelligent Logistics under Grants No. LDXX2017KF152, and Shandong Provincial Key Research and Development Program of China under Grants No. 2017CXGC0701.}
}


\author{Yijie Dang         \and
		Nan Jiang    \and
		Hao Hu 	\and
		Zhuoxiao Ji \and
		Wenyin Zhang
}


\institute{Y.J. Dang, N. Jiang, H. Hu and Z.X. Ji \at
              Faculty of Information Technology, Beijing University of Technology, Beijing, 100124, China \\
	Beijing Key Laboratory of Trusted Computing, Beijing, 100124, China \\
National Engineering Laboratory for Critical Technologies of Information Security Classified Protection, Beijing, 100124, China\\
              \email{jiangnan@bjut.edu.cn}        
           \and
           W.Y. Zhang \at
              School of Information Science and Technology, Linyi University, Linyi 276000, China
}

\date{Received: date / Accepted: date}

\maketitle

\begin{abstract}
Image classification is an important task in the field of machine learning and image processing. However, the usually used classification method --- the K Nearest-Neighbor algorithm has high complexity, because its two main processes: similarity computing and searching are time-consuming. Especially in the era of big data, the problem is prominent when the amount of images to be classified is large. In this paper, we try to use the powerful parallel computing ability of quantum computers to optimize the efficiency of image classification. The scheme is based on quantum K Nearest-Neighbor algorithm. Firstly, the feature vectors of images are extracted on classical computers. Then the feature vectors are inputted into a quantum superposition state, which is used to achieve parallel computing of similarity. Next, the quantum minimum search algorithm is used to speed up searching process for similarity. Finally, the image is classified by quantum measurement. The complexity of the quantum algorithm is only $O(\sqrt{kM})$, which is superior to the classical algorithms. Moreover, the measurement step is executed only once to ensure the validity of the scheme. The experimental results show that, the classification accuracy is $83.1\%$ on Graz-01 dataset and $78\%$ on Caltech-101 dataset, which is close to existing classical algorithms. Hence, our quantum scheme has a good classification performance while greatly improving the efficiency.
\end{abstract}

\section{Introduction}
\label{intro}
In 1982, Feynman proposed a novel computation model, named quantum computation. Due to the superposition and entanglement properties of the quantum states, this novel computation model can efficiently solve some problems that are believed to be intractable on classical computers\cite{1}. After that, many researchers devoted themselves to the research about quantum computation. In particular, in 1994, Shor designed a quantum integer factoring algorithm which can be done in polynomial time and the exponential speedup is achieved compared with the classical algorithm\cite{2}. In 1996, Grover's algorithm went quadratically faster than any possible classical algorithms\cite{3}. In addition, quantum image processing (QIP) is an important and rapidly developing sub-domain in quantum computation. Various quantum image representations have been proposed, such as flexible representation of quantum images\cite{4}, RGB multi-channel representation\cite{5}, novel enhanced quantum representation\cite{6}, generalized quantum image representation\cite{7} and Red-Green-Blue multi-channel quantum representation\cite{8}. Some quantum image processing algorithms have been developed based on these representation schemes, such as quantum image scrambling\cite{9,10}, quantum image steganography algorithm\cite{11,12}, quantum image matching\cite{13} , quantum binary images thinning algorithm\cite{14}, quantum watermarking\cite{15,16}, quantum image edge detection\cite{17}, quantum image motion detection\cite{18,19}, quantum image searching\cite{20}, quantum image metric\cite{21} and so on.

Image classification is an important task in the field of machine learning and image processing, which is widely used in many fields, such as computer vision, network image retrieval and military automation target identification. K Nearest-Neighbor (KNN) algorithm is one of the typical and simplest method to do image classification. KNN's basic idea is that if the majority of the $k$ nearest samples of an image in the feature space belong to a certain category, the image also belongs to this category. It has two core processes: similarity computing and $k$ nearest samples searching. Since KNN requires no learning and training phases and avoids overfitting of parameters, it has a good accuracy in dealing with classification tasks with more samples and less classes. Researchers propose many improved KNN algorithms\cite{22,23,24,25,26,27,28,29,30}. However, KNN and its improved algorithms are accompanied by a large amount of computation and have high complexity. In particular, the complexity of similarity computing process is $O(M)$ and the complexity of searching process is $O(M\log k)$, where $M$ is the number of training images. 

Recent progress implies that a crossover between machine learning and quantum information processing benefits both fields\cite{31,32,33,34}. Quantum mechanics offers tantalizing prospects to enhance machine learning, ranging from reduced computational complexity to improved generalization performance. The most notable examples include quantum enhanced algorithms for principal component analysis\cite{35}, quantum support vector machines\cite{36}, quantum Boltzmann machines\cite{37}, and so on\cite{38,39,40}. 
Ruan proposed a global quantum feature extraction method based on Schmidt decomposition, and also proposed a revised quantum learning algorithm that will classify images by computing the Hamming distance of these features\cite{41}. But the features and distances used by this algorithm limit the classification effect.
Chen proposed quantum K Nearest-Neighbor algorithm\cite{42} (QKNN) which realized KNN in quantum computers. QKNN uses quantum superposition states to achieve parallel computing of similarity and uses the quantum minimum search algorithm to speed up search process for similarity. Compared to classical algorithms, the complexity of similarity computing process is $O(1)$ and the complexity of searching process is $O(\sqrt{kM})$. Ref. \cite{43} realized another QKNN algorithm based on the metric of Hamming distance. Its complexity is $O(\sqrt{M}^3)$. Although Ref. \cite{42,43} show the process of QKNN and give the complexity analysis, they do experiments only on the image dataset of ten handwritten digits (0 to 9), instead of on natural image datasets.

In this paper, we propose an efficient natural image classification scheme based on QKNN. Firstly, the feature vectors of images are extracted. Then the feature vectors are stored in quantum superposition state which is used to achieve parallel computing of similarity. Next, the quantum minimum search algorithm is used to speed up searching process for similarity. Finally, the image is classified by quantum measurement. Moreover, the measurement step is executed only once to ensure the validity of the scheme. 

The rest of this paper is organized as follows. Firstly, our scheme is described. Then, complexity analysis and accuracy analysis are given respectively. Finally, we draw conclusions and outline possible future works.

\section{Image classification}

In this section, we show how to apply the QKNN algorithm to image classification and describe our solution.

\subsection{basic idea}

\begin{figure}[ht]
\centering
\includegraphics[width=10cm]{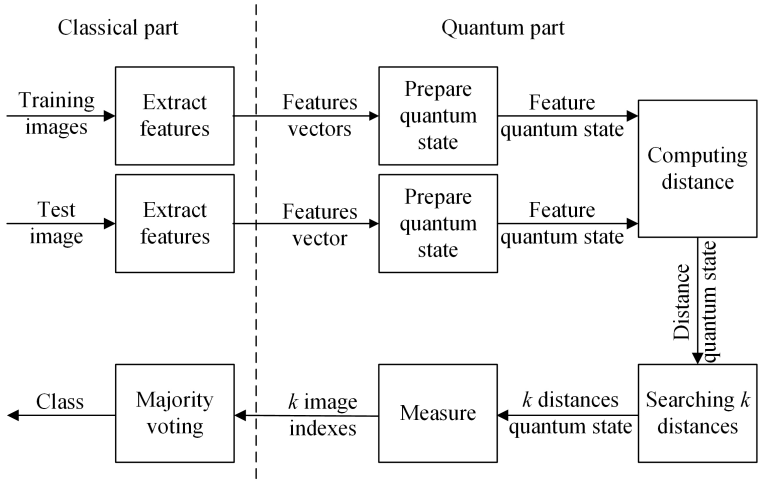}
\caption{Algorithm flow chart.}
\label{fig1}
\end{figure}

Image classification is the process that assigns an image to a corresponding class according to certain rules with a high degree of confidence. The basic ideas of our image classification scheme based on QKNN is shown in Fig. \ref{fig1}. Images that have been classified are training images and the unclassified image is the test image. The test image is classified according to training images.
Firstly, the feature vectors of all the images are extracted on a classical computer, which consists of color features and texture features. Then, the feature vectors are inputted into a quantum computer by applying a process for preparing quantum state. Next, distance between the test image and the training images, that is used to describe the similarity between them, is computed in parallel by a quantum circuit and is stored in the amplitude by applying the amplitude estimation algorithm. Then, applying D{\"u}rr's finding minimum algorithm (which will be described in detail in Step 4 in Section 2.2)\cite{44} to get the $k$ minimum distances from quantum superposition states\cite{32}. Finally, indexes of the $k$ similar images are obtained by measurement and the final classification result is produced by majority voting.

\subsection{Algorithm}

Assuming that all the images are RGB images. 
$V=\{V_0,V_1,\cdots,V_M\}$ is the complete dataset of the whole task, where $M$ is the number of training images.

Step 1 Extract features of images.

In this step, feature vectors (including color features and texture features) of the test image and the training images are extracted.
The process of feature extraction is divided into three sub-steps.

Step 1.1 Extract color features of images

$V_j$ is transformed from RGB (red, green, blue) color model to HSB model, where $H$ is the hue, $S$ is the saturation and $B$ is the brightness. Then according to the human visual characteristics, $H$ space is divided into 8 levels, $S$ space and $B$ space are divided into three levels respectively based on Eq. (1), where $h$, $s$ and $b$ are quantized values of $H$, $S$, and $B$.
\begin{equation}
h=\left\{
 \begin{aligned}
0, & H \in [0,20] \cup [316,359]\\
1, & H \in [21,40]\\ 
2, & H \in [41,75]\\
3, & H \in [75,155]\\
4, & H \in [156,190]\\
5, & H \in [191,270]\\
6, & H \in [271,295]\\
7, & H \in [296,315]
\end{aligned}
\right.
,
s=\left\{
\begin{aligned}
0, & S \in [0,0.2) \\
1, & S \in [0.2,0.7)\\
2, & S \in [0.7,1]
\end{aligned}
\right.
,
b=\left\{
 \begin{aligned}
0, & B \in [0,0.2) \\
1, & B \in [0.2,0.7)\\
2, & B \in [0.7,1]
\end{aligned}
\right.
\end{equation}

One dimensional feature vector is synthesized by the 3 components:
\begin{equation}
G=h*Q_h+s*Q_s+b
\end{equation}
where $Q_h$ and $Q_s$ are weights of $h$ and $s$, respectively. The larger the value of $Q_h$ and $Q_s$, the higher the accuracy of quantization, but the more the classification time. In the experiment of Ref. \cite{45}, the effect is better when $Q_h=9$ and $Q_s=3$. Thus, $G \in \{0,1,\cdots,71\}$.

Then by calculating histogram of $G$, color feature vector $c=({c_{1},c_{2},\cdots,c_{72}})$ is obtained. In order to eliminate difference in the size of images, $c$ is normalized. 

Step 1.2 Extract texture features of images.

$V_j$ is converted to a gray-scale image and the gray level co-occurrence matrix in four directions: $[0,1]$, $[-1,1]$, $[-1,0]$ and $[-1,-1]$, is computed. Then the mean and the variance of the contrast, correlation, energy and entropy compose the texture feature vector $t=({t_{1},t_{2},\cdots,t_{8}})$.

Since the physical meaning of each value is different and the size of the value varies, vector $t$ is normalized, which is the texture feature vector.

Step 1.3 Combine texture feature vector and color feature vector.

For the sake of simplicity, the color feature vector $c$ and texture feature vector $t$ are combined into a one-dimensional vector $(c_1,c_2,\cdots,c_{72},t_1,t_2,\cdots,t_8)$ which is denoted as $v_j=(v_{j1},v_{j2},\cdots,v_{ji},\cdots,v_{j80})$. Hence, $v_0$ is the feature vector of the test image and $v_j$ is the feature vector of the training image $V_j$ $(j\in \{ 1,2,\cdots,M\})$. 

Step 2 Store feature vectors to quantum state.

This step is used to store feature vectors $v_0$ and $v_j$ $(j\in \{1,2,\cdots,M\})$ into quantum states $\left|\alpha\right\rangle$ and $\left|\beta\right\rangle$ respectively\cite{46,47,48}.
\begin{equation}
\left|\alpha\right\rangle=\frac{1}{\sqrt N}\sum^N_{i=1}\left|i\right\rangle(\sqrt{1-v_{0i}^2}\left|0\right\rangle+v_{0i}\left|1\right\rangle)\left|0\right\rangle
\end{equation}
\begin{equation}
\left|\beta\right\rangle=\frac{1}{\sqrt M}\sum^M_{j=1}\left|j\right\rangle\frac{1}{\sqrt N}\sum^N_{i=1}\left|i\right\rangle\left|0\right\rangle(\sqrt{1-v_{ji}^2}\left|0\right\rangle+v_{ji}\left|1\right\rangle)
\end{equation}
where $N=80$ is the dimension of the feature vector.

The preparation consists of two main steps: prepare a number of initial qubits $|0\rangle$ and then store feature vectors. Firstly, we describe the preparation of $|\beta\rangle$. From Eqs. (3) and (4), it is obviously that the preparation of $|\alpha\rangle$ is similar to and easier than that of $|\beta\rangle$. The complete preparation process is as follows.

\begin{figure}[ht]
\centering
\includegraphics[width=7cm]{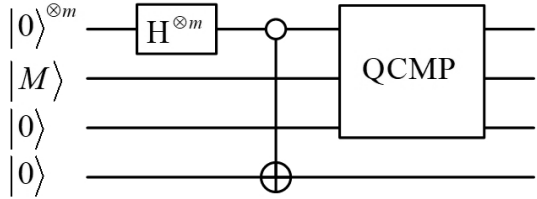}
\caption{Quantum circuit of preparing the initial superposition state $\frac{1}{\sqrt M}\sum^M_{j=1}\left|j\right\rangle$. }
\label{fig2}
\end{figure}

For $|\beta\rangle$, an initial superposition state $\frac{1}{\sqrt M}\sum^M_{j=1}\left|j\right\rangle$ is prepared by using $m$ Hadamard (H) gates on initial qubits $|0\rangle^{\otimes m}$ as shown in Fig. \ref{fig2}, where $m=\lceil \log_2 (M+1)\rceil$. Due to the fact that $M$ may not always be a power of 2, a judgement $U_1$ is needed:
\begin{equation}
U_1\left|j\right\rangle\left|M\right\rangle\left|0\right\rangle\left|0\right\rangle =
\begin{cases}
\left|j\right\rangle\left|M\right\rangle\left|0\right\rangle\left|1\right\rangle,&j=0 \\
\left|j\right\rangle\left|M\right\rangle\left|0\right\rangle\left|0\right\rangle,&0<j\leq M \\
\left|j\right\rangle\left|M\right\rangle\left|1\right\rangle\left|0\right\rangle,&j>M
\end{cases}
\end{equation}
where $\left|M\right\rangle$ stores number $M$ as a binary. $U_1$ is realized by QCMP (the Quantum Comparator\cite{49}) model as shown in Fig. \ref{fig2}. QCMP is used to determine $j$ is smaller or bigger than $M$, and the last two qubits $|0\rangle|0\rangle$ are two flag qubits to store the comparison results. Only if the two flags are in state $|0\rangle|0\rangle$, $|j\rangle$ satisfies $0<j\leq M$, i.e., $j$ is meaningful. The probability is $M/2^m$. For more details on QCMP, please refer to Ref. \cite{49}.


The initial superposition state $\frac{1}{\sqrt N}\sum^N_{i=1}\left|i\right\rangle$ is prepared by the same procedure as the preparation of $\frac{1}{\sqrt M}\sum^M_{j=1}\left|j\right\rangle$. After the above steps, initial quantum state $\left|\beta_0\right\rangle$ is prepared.
\begin{equation}
\left|\beta_0\right\rangle=\frac{1}{\sqrt M}\sum^M_{j=1}\left|j\right\rangle\frac{1}{\sqrt N}\sum^N_{i=1}\left|i\right\rangle\left|0\right\rangle\left|0\right\rangle
\end{equation}

Next, the feature value is set by using rotation matrix $R_y(2v_{kl})$ and controlled rotation gate $U_{kl}$.

\begin{equation}
R_y(2v_{kl})=\left(
\begin{array}{cc}
\cos v_{kl} & -\sin v_{kl} \\
\sin v_{kl} & \cos v_{kl} 
\end{array}
\right)
\end{equation}

\begin{equation}
U_{kl}\left|\beta_0\right\rangle=\frac{1}{\sqrt{MN}}\left(\left|0\right\rangle \otimes\left(\sum\limits^M_{j=1,j\neq k}\left|j\right\rangle\left\langle j\right| \right)\otimes\left(\sum\limits^N_{i=1,i\neq l}\left|i\right\rangle\left\langle i\right|\right)+R_y(v_{kl})\left|0\right\rangle \otimes \left|k\right\rangle\left|l\right\rangle \right) \left|0\right\rangle
\end{equation}

By defining $U_2=\prod^M_{j=1}\prod^N_{i=1}U_{ji}$, we can get 
\begin{equation}
\begin{split}
U_2\left|\beta_0\right\rangle
&=\prod\limits^M_{j=1}\prod\limits^N_{i=1}U_{ji}\left|\beta_0\right\rangle\\
&=\frac{1}{\sqrt{M}}\sum\limits^M_{j=1}\left|j\right\rangle\frac{1}{\sqrt{N}}\sum\limits^N_{i=1}\left|i\right\rangle\otimes(\cos v_{ji}\left|0\right\rangle+\sin v_{ji}\left|1\right\rangle)\left|0\right\rangle \\
\end{split}
\end{equation}
It is denoted as
\begin{equation}
\left|\beta_1\right\rangle=\frac{1}{\sqrt{M}}\sum\limits^M_{j=1}\left|j\right\rangle\frac{1}{\sqrt{N}}\sum\limits^N_{i=1}\left|i\right\rangle\left|v_{ji}\right\rangle\left|0\right\rangle
\end{equation}
and denote the inverse transformation of $U_2$ as $U^\dagger$.

Then by acting $R_y(2\sin^{-1}v_{ji})$ to the last qubit of $\left|\beta_1\right\rangle$, we can get
\begin{equation}
\left|\beta_2\right\rangle=\frac{1}{\sqrt{M}}\sum\limits^M_{j=1}\left|j\right\rangle\frac{1}{\sqrt{N}}\sum\limits^N_{i=1}\left|i\right\rangle\left|v_{ji}\right\rangle(\sqrt{1-v_{ji}^2}\left|0\right\rangle+v_{ji}\left|1\right\rangle)
\end{equation}
This process is denoted as $U_3$, where $R_y(2\sin^{-1}v_{ji})$ is a unitary operation which rotates around the $Y$ axis.
\begin{equation}
R_y(2\sin^{-1} v_{ji})=\left[
\begin{array}{cc}
\sqrt{1-v^2_{ji}} & -v_{ji} \\
v_{ji} & \sqrt{1-v^2_{ji}} \\
\end{array}
\right]
\end{equation}

Finally apply $U^\dagger$ to $\left|\beta_2\right\rangle$ to clear auxiliary qubit $\left|v_{ji}\right\rangle$ and call $U^\dagger$ as $U_4$.

The preparation of $\left|\alpha\right\rangle$ is similar to that of $\left|\beta\right\rangle$. Firstly, an initial superposition state $\frac{1}{\sqrt {N}}\sum^N_{i=1}\left|i\right\rangle$ is prepared using $n$ Hadamard (H) gates as shown in Fig. \ref{fig2}. Then apply $\prod^N_{i=1}U_{ji}$, $R_y(2\sin^{-1}v_{ji})$ and $U^\dagger$ to initial superposition state in turn where $k=0$ and $j=0$.

Step 3 Compute distances.

This step computes distances between the test image and the training images by applying controlled swap gate\cite{50,51}.

\begin{figure}[ht]
\centering
\includegraphics[width=5cm]{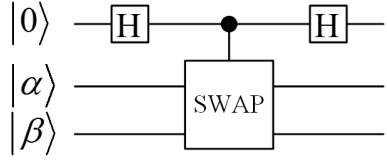}
\caption{Controlled swap gate.}
\label{fig3}
\end{figure}

Fig. \ref{fig3} shows the quantum circuit, in which controlled swap gate is used to implement operation: $\left|\alpha\right\rangle\left|\beta\right\rangle\to\left|\beta\right\rangle\left|\alpha\right\rangle$. Firstly, the auxiliary qubit $\left|0\right\rangle$ is mapped to $\frac{1}{\sqrt{2}}(\left|0\right\rangle+\left|1\right\rangle)$ by Hadamard gate. Then this superposition state acts as a control qubit for the swap gate to get $\frac{1}{\sqrt{2}}(\left|0\right\rangle\left|\alpha\right\rangle \left|\beta\right\rangle + \left|1\right\rangle \left|\beta\right\rangle \left|\alpha\right\rangle) $.
Finally, after the auxiliary qubit passes through H gate again, quantum state is changed to
\begin{equation}
\left|\varphi\right\rangle=\frac{1}{2}\left|0\right\rangle (\left|\alpha\right\rangle\left|\beta\right\rangle+\left|\beta\right\rangle\left|\alpha\right\rangle)+\frac{1}{2}\left|1\right\rangle (\left|\alpha\right\rangle\left|\beta\right\rangle-\left|\beta\right\rangle\left|\alpha\right\rangle)
\end{equation}

The probability of $\varphi=1$ is denoted as $P(1)$. Thus,
\begin{equation}
\begin{split}
P(1)
=&\left\langle\varphi\right|\left|1\right\rangle\left\langle1\right|\otimes I\left|\varphi\right\rangle \\
= &\left[ \frac{1}{2}\left\langle0\right| (\left\langle\alpha\right|\left\langle\beta\right|+\left\langle\beta\right|\left\langle\alpha\right|)+\frac{1}{2}\left\langle1\right| (\left\langle\alpha\right|\left\langle\beta\right|-\left\langle\beta\right|\left\langle\alpha\right|)\right]
\left|1\right\rangle\left\langle1\right| \\
& \otimes I \left[ \frac{1}{2}\left|0\right\rangle (\left|\alpha\right\rangle\left|\beta\right\rangle+\left|\beta\right\rangle\left|\alpha\right\rangle)+\frac{1}{2}\left|1\right\rangle (\left|\alpha\right\rangle\left|\beta\right\rangle-\left|\beta\right\rangle\left|\alpha\right\rangle)\right] \\
=& \left[\frac{1}{2}\left\langle1\right| (\left\langle\alpha\right|\left\langle\beta\right|-\left\langle\beta\right|\left\langle\alpha\right|)\right]\left[\frac{1}{2}\left\langle1\right| (\left|\alpha\right\rangle\left|\beta\right\rangle-\left|\beta\right\rangle\left|\alpha\right\rangle)\right] \\
=& \frac{1}{4}\left(\left\langle\alpha\right|\left\langle\beta\right|\left|\alpha\right\rangle\left|\beta\right\rangle+\left\langle\beta\right|\left\langle\alpha\right|\left|\beta\right\rangle\left|\alpha\right\rangle-\left\langle\alpha\right|\left\langle\beta\right|\left|\beta\right\rangle\left|\alpha\right\rangle-\left\langle\beta\right|\left\langle\alpha\right|\left|\alpha\right\rangle\left|\beta\right\rangle \right) \\
\end{split}
\end{equation}

Feature vectors is normalized so that $\left\langle \alpha | \alpha \right\rangle =1 $ and $\left\langle\beta|\beta\right\rangle=1 $. Hence,
\begin{equation}
\begin{split}
P(1)
=& \frac{1}{4}\left(1+1-\left\langle\alpha\right|\left\langle\beta\right|\left|\beta\right\rangle\left|\alpha\right\rangle-\left\langle\beta\right|\left\langle\alpha\right|\left|\alpha\right\rangle\left|\beta\right\rangle \right) \\
=& \frac{1}{4}\left(2-2\left|\left\langle\alpha\right| \beta\rangle^2\right| \right) \\
=& \frac{1}{2}-\frac{1}{2}\left|\left\langle\alpha\right| \beta\rangle^2\right| \\
\end{split}
\end{equation}
When $\left|\beta\right\rangle$ takes a particular state $v_j$, $P(1)=\frac{1}{2}-\frac{1}{2}\left|\left\langle v_0\right| v_j\rangle^2\right|$, where $\left|\left\langle v_0\right| v_j\rangle^2\right|$ is the inner product of vector $v_0$ and $v_j$. We use $P(1)$ to represent the similarity between vectors, and denote it as $d(v_0,v_j)$. Hence, the larger the inner product of the two vectors, the smaller $P(1)$ and the more smiliar the two images.

Thus, as shown in Fig. \ref{fig3}, by acting controlled swap gate on $|\alpha\rangle$ and $|\beta\rangle$, we can get

\begin{equation}
\left|\gamma\right\rangle=\frac{1}{\sqrt M}\sum\limits^M_{j=1}\left|j\right\rangle\left[\sqrt{1-d(v_0,v_j)}\left|0\right\rangle+\sqrt{d(v_0,v_i)}\left|1\right\rangle\right].
\end{equation}

Then we use Amplitude Estimation(AE)\cite{52} algorithm to transfer distance information to qubits. Due to the space limitations, please refer to Ref. \cite{52} to get the quantum circuit, the theory proof, and more details about AE. This process used $R$ iterations of Grover operator and the error is less than $\delta$, where $R$ and $\delta$ satisfy $R\geq\pi(\pi+1)/\delta$. 

AE helps us to get quantum state $\left|\sigma\right\rangle$ that stored the similarity information.
\begin{equation}
\left|\sigma\right\rangle=\frac{1}{\sqrt M}\sum\limits^M_{j=1}\left|j\right\rangle\left|d(v_0,v_j)\right\rangle
\end{equation}

Step 4 Search for $k$ minimum distances.

\begin{figure}[ht]
\centering
\includegraphics[width=10cm]{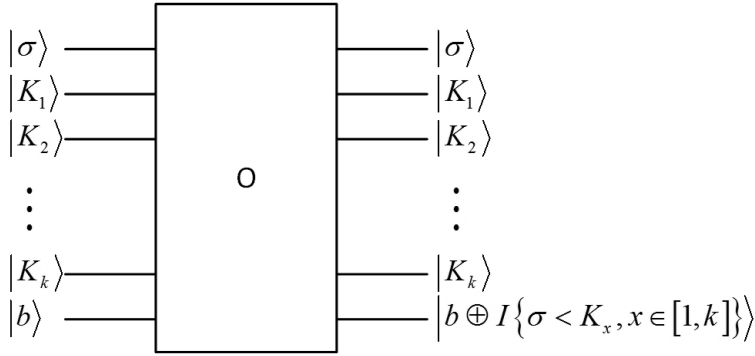}
\caption{Circuit model of search process.}
\label{fig4}
\end{figure}

In this step, we apply D{\"u}rr's algorithm\cite{44,53} to state $\left|\sigma\right\rangle$ to find the $k$ minimum. Ref. \cite{44} shows that this algorithm will return the $k$ minimum after $\sqrt{kM}$ iterations. Fig. \ref{fig4} shows the circuit model of this process. The searching process is as follows.

1) Define $K=\{K_1,K_2,\cdots,K_k\}$ as $k$ indexes of images that similar to $v_0$. Initially, the indexes $K_1,K_2,\cdots,K_k\in\{1,2,\cdots,M\}$ are selected randomly.

2) Use Grover's algorithm to find the quantum state which satisfies $d(v_0,v_j)<\max (d(v_0,v_{K_x}),x\in [1,k])$. By finite iterations of Grover, we can get an index $j$ which satisfies this condition. That is to say, the $j^{th}$ training image is more similar to the test image than the image with index ${K_x}$.

3) Set $j$ to $K_{max}$, where $K_{max}$ makes $d(v_0,v_{K_{max}})=\max(d(v_0,v_{K_x}), x\in [1,k])$. 

4) Repeat 2), 3) and interrupt it after $\sqrt{kM}$ iterations.

For each iteration, the index stored in qubits is replaced by an index with a smaller distance value. Finally, indexes of the $k$ images that are most similar to the test image are stored in auxiliary qubits $\left|K_1\right\rangle\left|K_2\right\rangle\cdots\left|K_k\right\rangle$.

Step 5 Measure and classify.

Measure $\left|K_1\right\rangle\left|K_2\right\rangle\cdots\left|K_k\right\rangle$ and get search results that are indexes of images which are similar to the test image. According to the basic idea of KNN algorithm, classify the test image to a class whose samples is the majority in results. If such a class is more than one, choose the first. In this step, the measurement step is executed only once.

\section{Complexity Analysis}

Our scheme consists of two processes: classical feature extraction and quantum classification based on QKNN, in which the latter is the main part. 

In classical part, the complexity of extracting color features is $O(3M)$, the complexity of extracting texture features is $O(4M)$ and the complexity of Combination of features is $O(M)$. Thus the complexity of Step 1 is $O(8M)$.

We focus on the complexity of the quantum part from the preparation of quantum feature vectors to searching for the $k$ minimum.

Some steps are based on Oracle. In order to discuss the complexity uniformly, we define Oracle as the basic unit. The complexity of each step is analyzed as follows.

Step 2: preparation process of $\left|\alpha\right\rangle$ and $\left|\beta\right\rangle$ includes $U_1$, $U_2$, $U_3$ and $U_4$. $U_1$ uses some H gates in parallel, one CNOT gate and one quantum comparator. Hence, its complexity is 3. $U_2$, $U_3$ and $U_4$ are regarded as 3 Oracles. This step prepares two quantum states so that it needs 6 Oracles. Hence the time complexity is $O(1)$, which is a constant.

Step 3: computing distance requires one controlled swap gate and its complexity is $O(1)$. Ref. \cite{52} indicates AE need $R$ iterations of Grover operator and a Grover operator needs 12 Oracles. Thus the complexity of this process is $O(12R)$. 

Step 4: according to the previous description and the conclusion of Ref. \cite{44,53}, the complexity of this step is $O(\sqrt{kM})$.

So the total complexity of the quantum part is 
\begin{equation}
O(1)+O(12R)+O(\sqrt{kM}) \approx O(\sqrt{kM})
\end{equation}

Established classical image classification algorithm also has the feature extraction step, which is the same as Step 1 of our scheme. That is to say, their complexities are the same. Hence, in the following, we compare the quantum algorithm and the classical algorithm only from the two core processes: similarity computing and searching.
\begin{enumerate}[1.]

\item If the effect of the dimension of the feature vector is not taken into account and view a computation distance as a unit, the complexity of the classical similarity computation is $O(M)$. Correspondingly, quantum algorithm only uses a controlled swap gate, whose complexity is a constant $O(1)$. It shows that our algorithm achieves the acceleration from linear complexity to constant complexity.

\item The complexity of searching based on the sorting algorithm is not less than $O(M\log k)$ on classical computers. However the complexity of quantum search process is $O(\sqrt{kM})$. 
\end{enumerate}

\begin{figure}[ht]
\centering
\includegraphics[width=12cm]{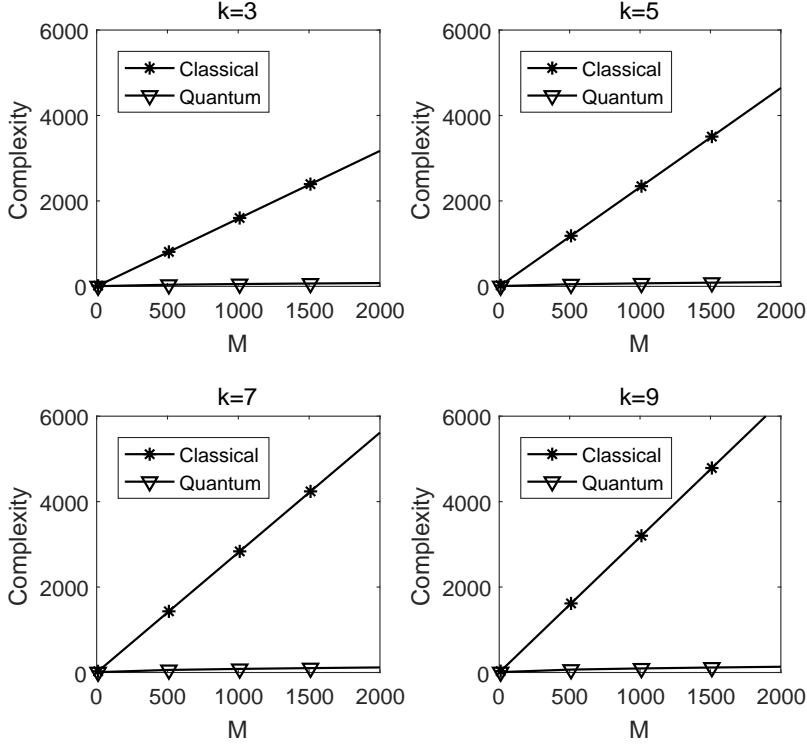}
\caption{Complexity comparison between classical and quantum algorithms in search process. $M$ is the number of training images. }
\label{fig5}
\end{figure}

We make a detailed comparison between classical search complexity $O(M\log k)$ and quantum search complexity $O(\sqrt{kM})$. Considering the effect of $k$ on the experimental accuracy, $k \in \{3,5,7,9\}$. The result is shown in Fig. \ref{fig5}. Whatever $k$ takes, the complexity of classical search process is higher than the quantum one. Moreover, the larger the $M$, the greater the quantum advantage. Therefore, quantum search algorithm achieved speedup.

In addition, quantum KNN has two more processes than KNN, which are quantum state preparation and AE. These two processes has only constant complexity. Hence, it's not taken into account.

To sum up, our algorithm significantly reduces the complexity of image classification based on KNN algorithm.

\section{Simulation-based experiments}

Section 4.1 gives a simple experiment with 10 images to further show the details of our algorithm. Section 4.2 and 4.3 use two widely used image sets --- Graz-01 and Caltech-101 to demonstrate the accuracy of the algorithm, which have 833 images in 2 classes and 2921 images in 9 classes respectively.

\subsection{A simple experiment}

\begin{figure}[ht]
\centering
\includegraphics[width=12cm]{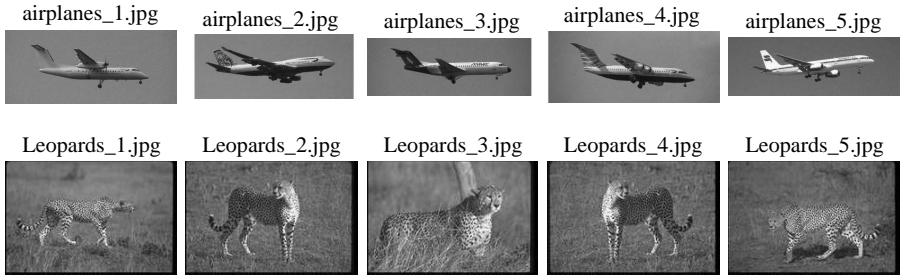}
\caption{Training images }
\label{fig6}
\end{figure}

\begin{figure}[ht]
\centering
\includegraphics[width=6cm]{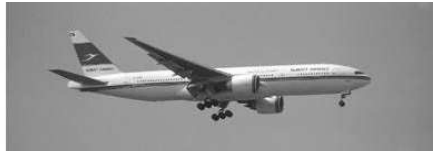}
\caption{Test image }
\label{fig7}
\end{figure}

Caltech-101 is a data set of digital images that is intended to facilitate computer vision research and techniques. Moreover, it is most applicable to techniques involving image recognition classification and categorization\cite{54}. Ten RGB images which include five images from the airplanes class and five images from the Leopards class make up the training dataset as shown in Fig. \ref{fig6}. The test image is shown in Fig. \ref{fig7}.


Firstly, the feature vectors of all the images are extracted according to Step 1. Considering the high dimensionality of the vectors (80), for brevity, only four variables are shown in Table \ref{tab1}.

 \begin{table}[ht]
\small
\centering
\begin{tabular}{|l|l|l|l|l|l}
\hline
Image &*& $v_{*1}$ & $v_{*2}$ & $v_{*24}$ & $v_{*78}$ \\
\hline
Test image &0& 0.00002633 & 0.00078124 & 0 & 0.09610999 \\
\hline
airplanes\_1 &1& 0.00008357 &0.00005014 &0 & 0.11966605 \\
\hline
airplanes\_2 &2& 0.00492092 & 0.00180055 & 0 & 0 \\
\hline
airplanes\_3 &3& 0.00196011 & 0.00013930 & 0 &0.06067670 \\
\hline
airplanes\_4 &4& 0.00199697 & 0.00004851 & 0 & 0.09119511 \\
\hline
airplanes\_5 &5& 0.00000771 &0.00006166 & 0 &0.18334765 \\
\hline
Leopards\_1 & 6&0.00085182 &0 &0.00518836 &0.59927777 \\
\hline
Leopards\_2 & 7&0.00110141 & 0 & 0.01474394 & 0.57039998 \\
\hline
Leopards\_3 &8& 0.00043356 & 0 &0.01250811 & 0.53275436 \\
\hline
Leopards\_4 &9& 0.00127659 &0 &0.01514383 & 0.57050397 \\
\hline
Leopards\_5 &10& 0.00089379 &0 &0.01117236 & 0.56351601 \\
\hline
\end{tabular}
\caption{\label{tab1}Representative part of the feature vectors.}
\end{table}

In Step 2, feature vectors of the 10 training images are stored in the quantum state $\left|\beta\right\rangle$ and feature vector of the test image is stored in the quantum state $\left|\alpha\right\rangle$. Since for each image, its feature vector's dimension is 80 and there are 10 images, i.e., $N=80$ and $M=10$. That is to say, in this example 
$
\left|\alpha\right\rangle=\frac{1}{\sqrt{ 80}}\sum^{80}_{i=1}\left|i\right\rangle(\sqrt{1-v_{0i}^2}\left|0\right\rangle+v_{0i}\left|1\right\rangle)\left|0\right\rangle
$ and
$
\left|\beta\right\rangle=\frac{1}{\sqrt {10}}\sum^{10}_{j=1}\left|j\right\rangle\frac{1}{\sqrt{ 80}}\sum^{80}_{i=1}\left|i\right\rangle\left|0\right\rangle\\ \left(\sqrt{1-v_{ji}^2}\left|0\right\rangle+v_{ji}\left|1\right\rangle\right)
$ by substituting $N$ and $M$ into Eq. (3) and Eq. (4).

 $\left|\alpha\right\rangle$ and $\left|\beta\right\rangle$ are the inputs of controlled swap gate in Step 3 and we can get distances between the test image and the training images which are stored in $\left|\gamma\right\rangle$. In Eq. (19), $d(v_0,v_j)$ is the distance between the test image and the $j^{th}$ training image. Actual data is shown in Table \ref{tab2}. 

\begin{equation}
\left|\gamma\right\rangle=\frac{1}{\sqrt {10}}\sum\limits^{10}_{j=1}\left|j\right\rangle\left[\sqrt{1-d(v_0,v_j)}\left|0\right\rangle+\sqrt{d(v_0,v_i)}\left|1\right\rangle\right]
\end{equation}

Distance values are stored in quantum state $\left|\sigma\right\rangle$ by acting AE on $|\gamma\rangle$. 

\begin{equation}
\left|\sigma\right\rangle=\frac{1}{\sqrt {10}}\sum\limits^{10}_{j=1}\left|j\right\rangle\left|d(v_0,v_j)\right\rangle
\end{equation}

 \begin{table}[ht]
\small
\centering
\begin{tabular}{|l|l|l|l|l|l}
\hline
Image & Ranking & $d(v_0,v_*)$ \\
\hline
airplanes\_1 &2& 0.0349247032450856 \\
\hline
airplanes\_2 &1& 0.0227572679900773 \\
\hline
airplanes\_3 & 3& 0.0524175219266468 \\
\hline
airplanes\_4 &4& 0.0896693757206786\\
\hline
airplanes\_5 &5& 0.126960495310553\\
\hline
Leopards\_1 & 9& 0.478449921552929 \\
\hline
Leopards\_2 &8& 0.474949680247898 \\
\hline
Leopards\_3 &6&0.470301163393825 \\
\hline
Leopards\_4 &7& 0.474904076167589 \\
\hline
Leopards\_5 &10& 0.485184863755358 \\
\hline
\end{tabular}
\caption{\label{tab2}Distances between the test image and the training images.}
\end{table}

In Step 4, by supposing $k=3$, the initial quantum state is $\left|\sigma\right\rangle\left|0000\right\rangle\left|0000\right\rangle\left|0000\right\rangle\left|b\right\rangle$. After $\lceil\sqrt{30}\rceil=6$ iterations of D{\"u}rr's algorithm, the $k$ smallest distances are searched and are stored in auxiliary qubits. Thus we can get $\left|\sigma\right\rangle\left|0010\right\rangle\left|0001\right\rangle\left|0011\right\rangle\left|b\right\rangle$. 

Through measuring, results are binary string 0010, 0001 and 0011, i.e., decimal number 2, 1 and 3 respectively, which are exactly indexes of the $k$ smallest distances in Table \ref{tab2}. Thus, the training image is similar to the test image airplanes\_2.jps, airplanes\_1.jps and airplanes\_3.jps. Since the three training images are belonged to airplanes class, the test image is also belonged to airplanes class.

\subsection{Experiment on Graz-01 dataset}

\begin{figure}[ht]
\centering
\includegraphics[width=12cm]{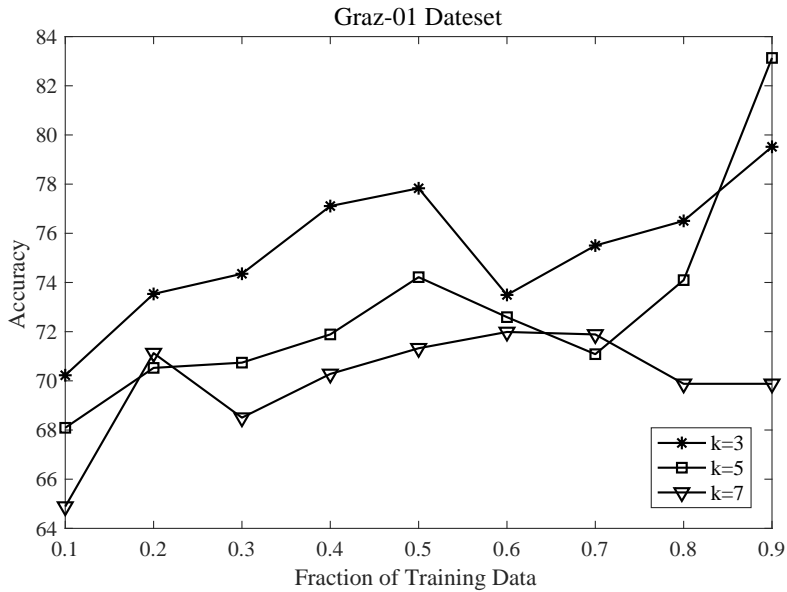}
\caption{Performance on the Graz-01 dataset.}
\label{fig8}
\end{figure}

The Graz-01 database\cite{55} has two object-classes (bikes and persons), and one background-class. It is widely used in image classification tasks to compare the accuracy and effectiveness of different methods. We apply our algorithm to this database and demonstrate the performance of our algorithm. Due to characteristics of KNN, we remove the background class of the database. Thus experiment dataset contains two object-classes (bikes and persons) which has total 833 images. Simulation experiment data is shown in Fig. \ref{fig8}.

In the experiments, the maximum value of accuracy is 83.1\% when $k=5$ and training 90\% data. When $k=3$ the average value reaches 75.3\% and it is not less than 70\% even if the training data is relatively less. It indicates that the algorithm has advantages even if training data is less.

We compare the accuracy of our scheme with the experimental data in Ref. \cite{29,30} because their hypothetical conditions are close. Table \ref{tab3} shows that the accuracy of Bikes in our algorithm is lower than that of the other two algorithms and the accuracy of People is higher than the others. Overall, the accuracy of our algorithm is almost the same as other algoritms and is acceptable. It shows that our algorithm improve efficiency while ensuring an acceptable accuracy.

\begin{table}[ht]
\centering
\begin{tabular}{|l|l|l|l|l}
\hline
Class & QKNN & Opelt\cite{29} & Lazebnik\cite{30} \\
\hline
Bikes & 77 & 86.5 & 86.3 \\
\hline
Peoples & 84.8 & 80.8 & 82.3 \\
\hline
\end{tabular}
\caption{\label{tab3}Comparison of the accuracy of different algorithms.}
\end{table}

\subsection{Experiment on Caltech-101 dataset}

\begin{figure}[ht]
\centering
\includegraphics[width=12cm]{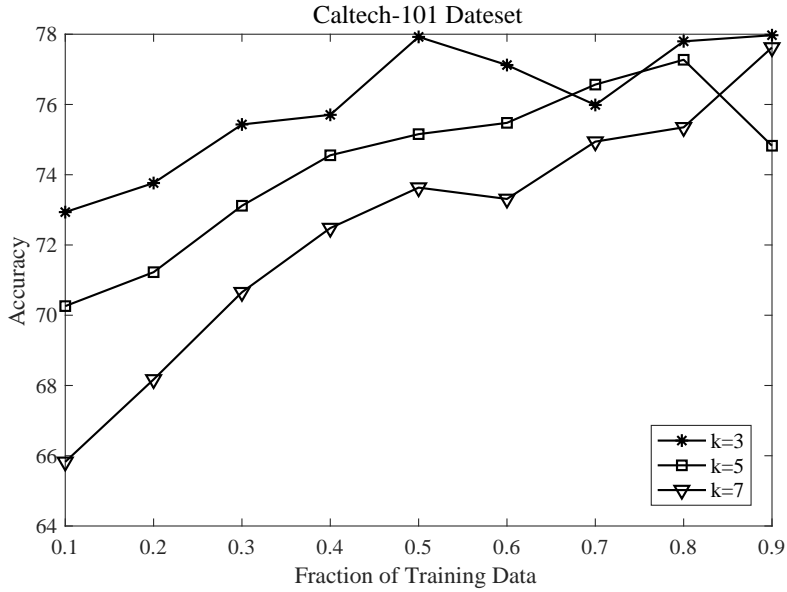}
\caption{Performance on the Caltech-101 dataset.}
\label{fig9}
\end{figure}

Caltech-101 has 101 classes (animals, furniture, flowers, and etc.)\cite{54}. Because KNN is suitable for classification tasks with more samples and less classes, we choose 9 classes which have more than 100 samples. Hence our dataset contains 9 classes (airplanes, bonsai, chandelier, etc.) with 2921 images. Fig. \ref{fig9} shows the results, in which, the accuracy reaches 78\% when $k=3$ and the training ratio that is the proportion of training data to all data is 90\%. Similar to the previous experiment, the accuracy is increased with the increment of training proportion. 

Since quantum machine learning is new research field, and the existing QKNN researches do not do experiments on natural images, there is lack of comparison data in quantum field. Hence, our experimental data is compared only with that of classical algorithms.
Through the simulation experiments on Graz-01 and Caltech-101 datasets, the quantum scheme maintains good accuracy while greatly improving the efficiency. And quantum scheme has an acceptable accuracy even if the training ratio is low.

\section{Conclusions}
This paper uses the powerful parallel computing ability of quantum computers to optimize the efficiency of image classification. The scheme is based on quantum KNN. Firstly, the feature vectors of images are extracted on classical computers. Then the feature vectors are inputted into quantum superposition state, which is used to achieve parallel computing of similarity. Next, the quantum minimum search algorithm is used to speed up searching process for similarity. Finally, the image is classified by quantum measurement. The complexity of the quantum algorithm is only $O(\sqrt{kM})$, which is superior to the classical algorithms. The measurement step is executed only once to ensure the validity of the scheme. Moreover, our quantum scheme has a good classification performance while greatly improving the efficiency.

There are some things that can be done as future works. Firstly, QKNN is not the only solution for image classification. Image classification algorithms based on more progressive and more subtle machine learning algorithms should be studied to improve the accuracy and application range in the future. Secondly, the method of feature extraction in our scheme is relatively simple and it is done in classical computers. Hence, optimizing feature extraction algorithms and designing quantum implementation solutions is one of the future work.




\end{document}